\documentclass{llncs}
\usepackage[]{amsmath}%
\usepackage{hyperref}
\usepackage{graphicx}
\usepackage{color}
\usepackage{caption}
\usepackage{subcaption}
\begin{document}
\title{Deep-FExt: Deep Feature Extraction for Vessel Segmentation and Centerline Prediction}

\author{Giles Tetteh\inst{2}, Markus  Rempfler\inst{2}, Bjoern H. Menze\inst{2}, Claus Zimmer\inst{1}
\institute{Neuroradiology, Klinikum Rechts der Isar, TU M\"unchen, Munich, Germany
\and Department of Computer Science, TU M\"unchen, Munich, Germany}
\email{giles.tetteh@tum.de}}

%\author{**}

%\institute{**}

\maketitle

\begin{abstract}
Feature extraction is a very crucial task in image and pixel (voxel) classification and regression in biomedical image modeling. In this work we present a machine learning based feature extraction scheme based on inception models for pixel classification tasks. We extract features under multi-scale and multi-layer schemes through convolutional operators. Layers of Fully Convolutional Network are later stacked on this feature extraction layers and trained end-to-end for the purpose of classification. We test our model on the DRIVE and STARE public data sets for the purpose of segmentation and centerline detection and it out performs most existing hand crafted or deterministic feature schemes found in literature. We achieve an average maximum Dice of 0.85 on the DRIVE data set which out performs the scores from the second human annotator of this data set. We also achieve an average maximum Dice of 0.85 and kappa of 0.84 on the STARE data set. Though these datasets are mainly 2-D we also propose ways of extending this feature extraction scheme to handle 3-D datasets.  
\end{abstract}

\section{Introduction}
% 1. Related prior work
Most recent research in biomedical modelling involves qualitative and quantitative classification of a single pixel (voxel), a region of interest ROI and or an image (volume). These classification tasks mostly involve three main steps - feature extraction, feature selection and classification \cite{Ryszard}. Out of these three steps, the feature extraction step is the most crucial since it goes a long way to determine how successful the classification task would be. 

Feature  extraction  is  the  process of generating features to be used in the selection and classification tasks\cite{Ryszard}. In whole image or volume classification, feature extraction and selection can serve as a dimensionality reduction task where a subset of the extracted features is selected to eliminate redundant features while maintaining the underlying discriminatory information\cite{Adegoke}. The newly extracted features normally are of lower dimension than the original feature space. However, most pixelwise feature extraction task leads to dimensionality extension. That is, a new set of features of high dimension is extracted from a given pixel and its neighbourhood. 

Feature extraction techniques come mainly in three main flavours - hand crafted texture features, supervised learned features and unsupervised feature extraction.

Textures are complex visual patterns composed of entities, or subpatterns, that have characteristic brightness, colour,
slope, size, etc \cite{TextureFeat}. The local subpattern properties give rise to the perceived lightness, uniformity, density, roughness, regularity, linearity, frequency, phase, directionality, coarseness, randomness, fineness, smoothness, granulation, etc., of the texture as a whole \cite{levine}. For a review of texture features, categorization and various uses one can refer to \cite{TextureFeat}.

Other groups of hand crafted features are based on differential geometry and the analysis of gradient and Hessian of pixel intensity. These are mostly used as image enhancement to objects of specific shape of interest in a given image. For example in \cite{Frangi} The multiscale second order local structure of an image (Hessian) is examined with the purpose of developing a vessel enhancement filter and ultimately a vesselness measure is obtained on the basis of the eigenvalues of the Hessian. This vesselness measure serves as a measure of the likelihood of the presence of geometrical structures which can be regarded as tubular. Also a curvilinear structure detector, called Optimally Oriented Flux (OOF) is proposed by \cite{OOF} for vesselness enhancement. OOF finds an optimal axis on which image gradients are projected in order to compute the image gradient flux\cite{OOF}.

The second class of feature extraction techniques are in the form of unsupervised learning and transfer learning. These are mainly autoencoders and its variations like Restricted Boltzmann's Machine. Autoencoders are simple learning circuits which aim to transform inputs into outputs with the least possible amount of distortion \cite{Autoencoders}. For detailed discussion of Autoencoders, Unsupervised learning and deep architectures one can refer to \cite{Autoencoders}. These architectures though very simple are very important in the field of machine learning and form the basic component of deep learning architectures.

Architectures like CNN and our deep networks also extract hierarchical features in a supervised manner through the use of ground truth annotations. Szegedy, C et al. \cite{Szegedy1} proposed the inceptions model as a way of building deeper networks capable of learning and extracting dense feature while maintaining acceptable speed and memory usage. This idea has been used in building the GoogLeNET which achieves the state of the art results on image classification task. 

In this paper we discuss briefly inception models in general and extend the idea to build feature extraction layers based on these inception models in an autoencoder fashion. We will also look at how to stack these pixelwise features extraction layers to form a deep architecture which is then fine tuned for the purpose of supervised learning.

% 2. Contributions

\section{Methodology}% Proposed method
\subsection{Inception Models}\label{sec:inception}
The main idea of the Inception architecture is based on finding out how an optimal local sparse structure in a convolutional vision network can be approximated and covered by readily available dense components \cite{Szegedy1}.
Inception based networks replaces convolution operations with mini-networks which uses less parameters and less computation. A convolution with a filter size of $5 \times 5$ can be replaced with a mini-network of two layers of filter sizes $3 \times 3$ each as shown in figure \ref{mini-net}. This reduces the parameter size from 25 (i.e. $5 \times 5$) to 18 (i.e. $3 \times 3 + 3 \times 3$  Similarly a comvolution operation with filter size $3 \times 3$ can be replaced with a mini-network of two layers with filters $1 \times 3 $ and $3 \times 1$ respectively as shown in figure \ref{mini-net}.
\begin{figure}
\centering
\includegraphics[scale=0.2]{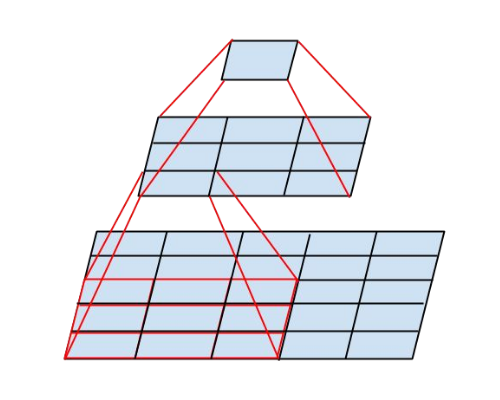}
\includegraphics[scale=0.2]{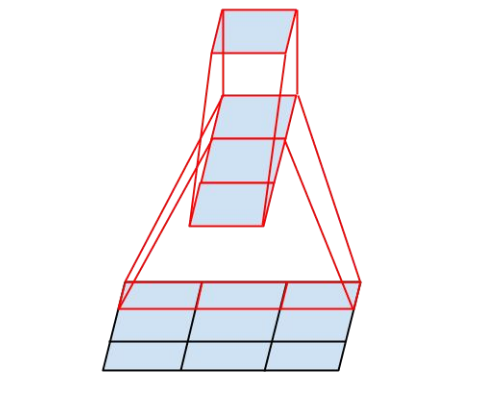}
\caption{Left: Mini-network replacing a 5x5 convolution operation, Right: Mini-network replacing a 3x3 network}\label{mini-net}
\end{figure}

\begin{figure}
\includegraphics[scale=0.2]{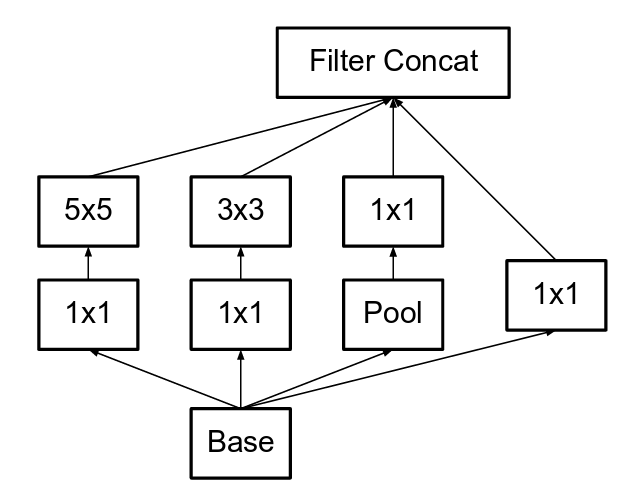}
\includegraphics[scale=0.2]{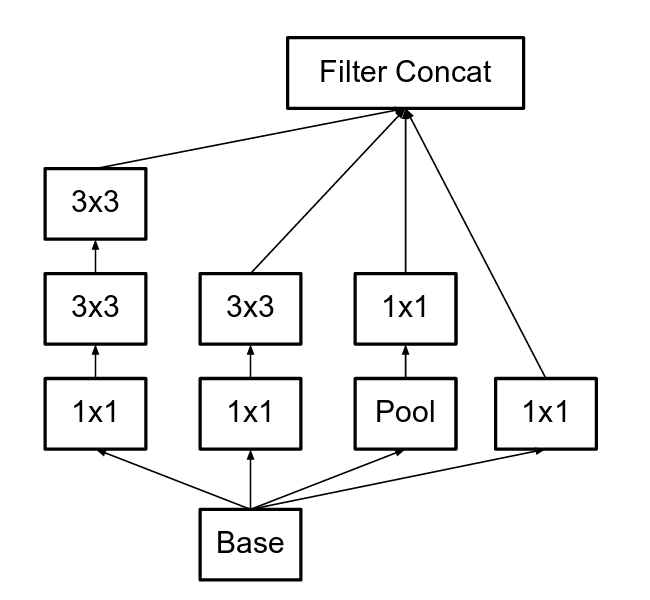}
\caption{Left: Inception model as described in \cite{Szegedy1}, Right: factorization proposed in \cite{Szegedy2}}\label{mini-net}
\end{figure}

By factorizing convolution operations with bigger filter sizes into mini-networks with smaller filter sizes \cite{Szegedy2} proposed building a network which make use of filters with sizes not greater than $3 \times 3$. This helps to conserve memory and computational time which can be used to increase the depth of the network to improve performance. Inception modules as described in \cite{Szegedy1,Szegedy2} form the building layers of the state of the art GoogLeNet network which was presented to the  ILSVRC14 competition. Thorough discussion of inception architure can be found in \cite{Szegedy1,Szegedy2}.  The original inception module proposed is used in networks meant for full image classification. In section \ref{pwfel} we consider ways in adapting the inception module to form a feature extraction layer in pixel wise classification tasks.

\subsection{Pixel wise feature extraction layer}\label{pwfel}
\begin{figure}[t!]
\centering
    \begin{subfigure}[b]{0.5\textwidth}
            \centering
            \includegraphics[scale=0.2]{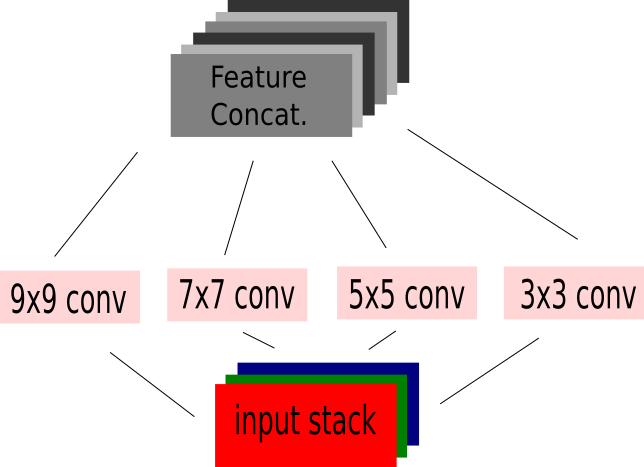}
            \caption{i}
             \label{sfig:a}
    \end{subfigure}%
    \begin{subfigure}[b]{0.5\textwidth}
            \centering
            \includegraphics[scale=0.2]{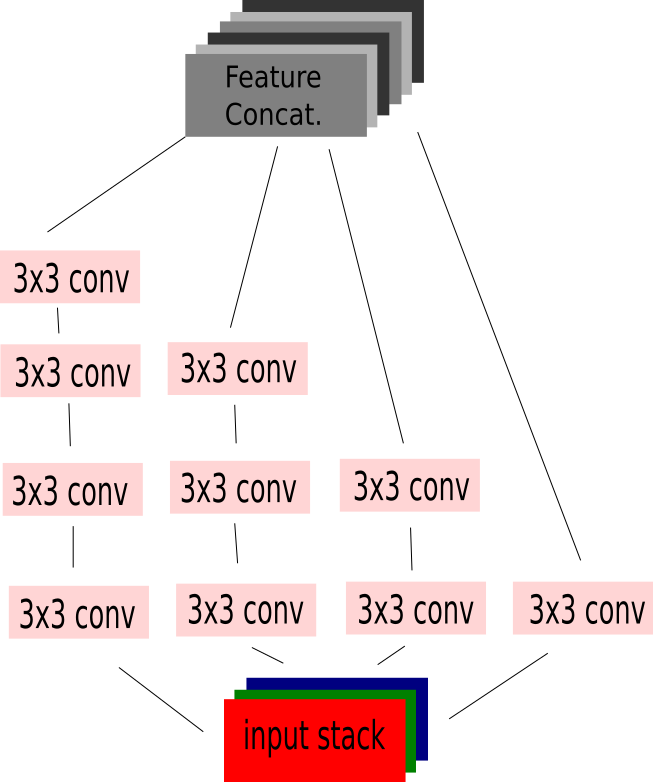}
            \caption{ii}
             \label{sfig:b}
    \end{subfigure}
    \caption{Feature extraction layers;(\subref{sfig:a}) layer without network factorization  and (\subref{sfig:b}) layer after network factorization.}
\end{figure}

\begin{figure}[t!]
\centering
    \includegraphics[scale=0.4]{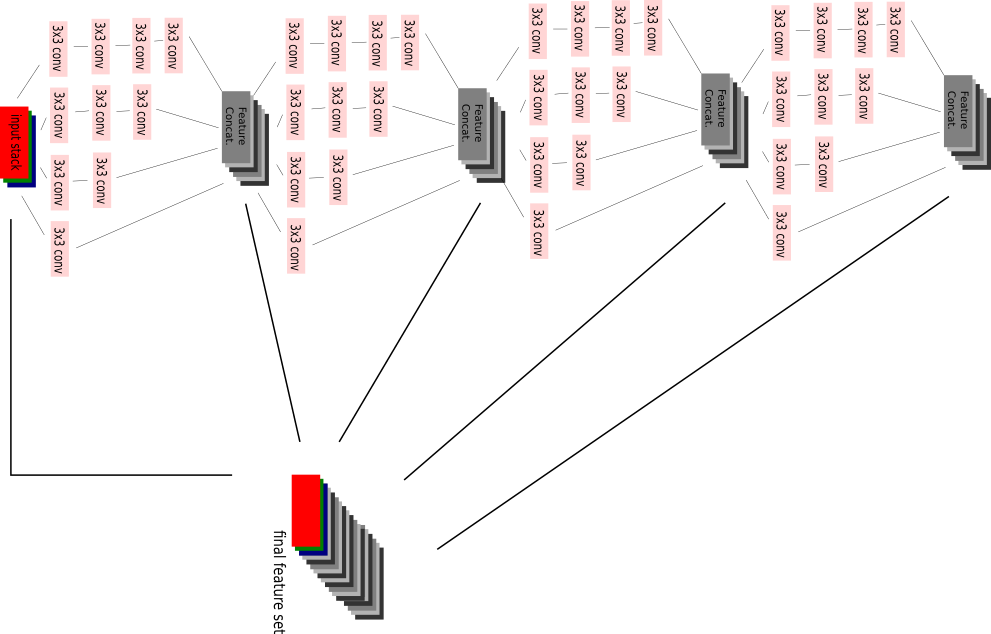}
    \caption{Feature extraction network with final feature set from multiple layers}
	\label{featureset}
\end{figure}

The original inception architecture described in section \ref{sec:inception} is designed to fit in the domain of full image classification. This therefore leads to feature or dimensionality reduction. However, in this section we are rather interested in extracting features for pixel classification. The first two steps is to 
\begin{enumerate}
\item remove all pooling operations.
\item all convolution operations result in an output of the same size as the input.
\end{enumerate}
we then choose a set of scales $s$ e.g. $\{3,5,7,9\}$ design a multiscale layer as shown in figure \ref{sfig:a}. We then replace all convolutions with bigger than $3 \times 3$ filter sizes with mini-networks as described in section \ref{sec:inception} and depicted in figure \ref{mini-net} to obtain our final feature extraction layer in figure \ref{sfig:b}. By stacking multiple layers together we build a feature extraction network suitable for pixel classification. We note that the output from each layer is further transformed by a non-linear activation function before it moves to the next layer. The concatenated output from each layer together with the input image are further concatenated to form the final feature set as shown in figure \ref{featureset}. 

\section{Experiments} \label{sec:exp}
\begin{figure}[t!] 
\centering
\includegraphics[scale=0.7]{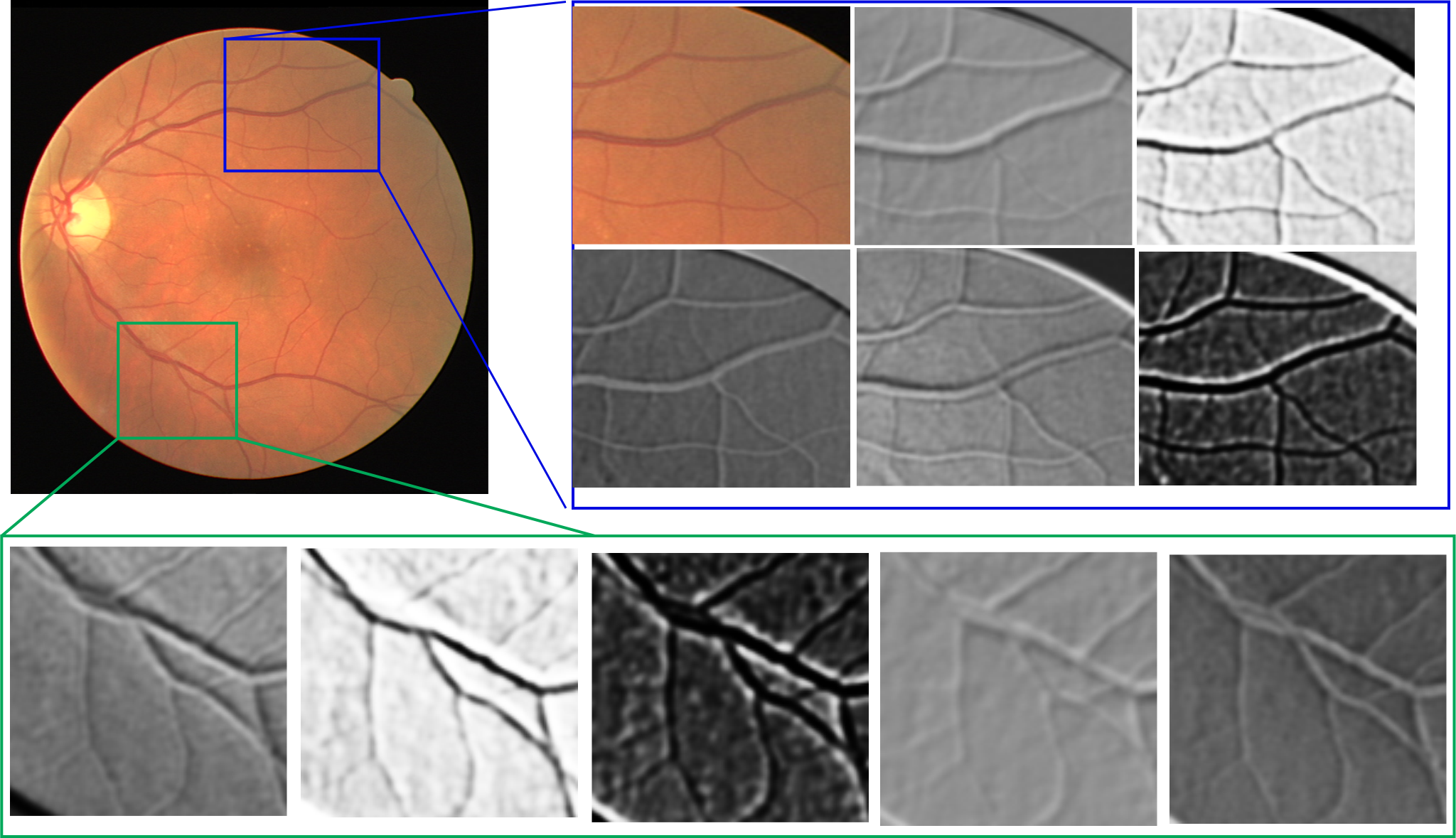}
\caption{Actual image (top left) and sample of features extracted by Deep-FExt. (More examples of extracted features in supplementary)}
\label{fig:features}
\end{figure}

To test Deep-FExt we design a network of 5 feature extraction layers which extract a total of 100 features per pixel (see fig. \ref{featnetstructure}). we then create a $10 \times 10$ feature mesh from each pixel feature set. Hence each pixel is then represented by 2-D image of size $10 \times 10$. A CNN with 3 layers is then stacked on the feature mesh and trained end-to-end for classification and prediction purposes. The full network structure is described in Table \ref{featureset}.  
\begin{table}
\begin{tabular}{|l|l|l|l|}
\hline
layer & input type and size & filter sizes (extracted feat.) & total features \\
\hline
1 & RGB image with 3 channels & 3(5),5(5),7(5),9(3),11(3) & 21 \\
2 & concat features from layer 1 & 3(5),5(5),7(5),9(3),11(3) & 21 \\
3 & concat features from layer 2 & 3(5),5(4),7(4),9(3),11(3) &19 \\
4 & concat features from layer 3 & 3(4),5(4),7(4),9(3),11(3) &18 \\
5 & concat features from layer 4 & 3(4),5(4),7(4),9(3),11(3) &18 \\
\hline
& Total & & 97 + 3 (input RGB) = 100 \\
\hline
\end{tabular}
\caption{Feature extraction network structure employed in our experiment.}
\label{featnetstructure}
\end{table}
Our experiments show that Deep-FExt is able to extract hierarchical features ranging from edge detectors, intensity gradients, and curvature at different scales (see. sample feature in figure \ref{fig:features}).

\subsection{Vessel Segmentation}
For vessel segmentation we experiment on the DRIVE \cite{drive} and STARE \cite{stare} datasets. The DRIVE dataset is made up of 20 training set and 20 test set with two annotations in each group. We use the first annotation as the ground truth for training our network and testing. We also compare our results to the second annotation. The STARE dataset is made up of  20 annotated images with two annotations each. We split the data into 10 images for training and the remaining 10 for testing. Our results (See Tables \ref{tab:ds-result},\ref{tab:ss-result}) shows that our Deep-FExt network our performs most of the existing architecture for the segmentation of on the DRIVE and STARE dataset. Results for Deep Retinal Understanding (DRIU) \cite{driu} are obtained by evaluating pre-computed probability maps provided on the paper's page. Other results are also stated as reported by \cite{driu}.

\begin{figure}[t!] 
\centering
\includegraphics[scale=0.6]{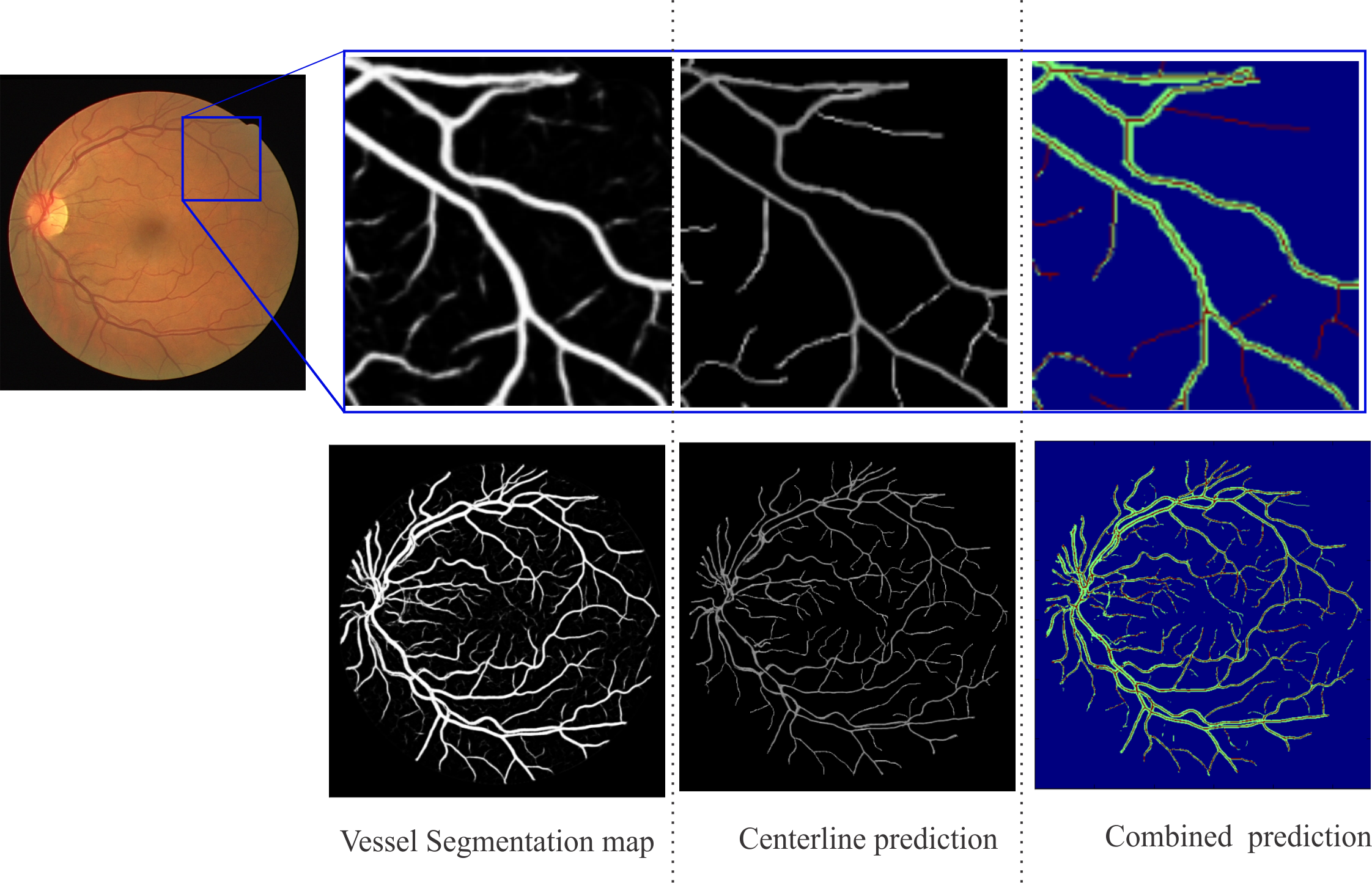}
\caption{Qualitative predictions from Deep-FExt. More images in supplementary}
\label{fig:drive-result}
\end{figure}

\begin{table}[t!]
\centering
\begin{tabular}{l|l|l|l|l|l}
\hline
Method & Precision & Recall & F1 Score & Average Max. Dice & Kappa \\
\hline
Deep-FExt & 0.8044 & 0.8032 & 0.8038 & 0.8467 & 0.7848 \\
Deep-FExt (Fusion) & \textbf{0.8215} & \textbf{0.8988} & \textbf{0.8584} & \textbf{0.8904} & \textbf{0.8439} \\
DRIU & 0.8159 & 0.8261 & 0.8210 & 0.8602 & 0.8034 \\
$N^4$ fields & & & 0.805 & & \\ 
Kernel Boost & & & 0.800 & & \\
HED & & & 0.796 & & \\
CRFs & & & 0.781 & & \\
Wavelets & & & 0.762 & & \\
Line Detectors & & & 0.692 & & \\
SE & & & 0.658 & & \\
Second Annotator & 0.8040 & 0.7746 & 0.7890 & 0.8298 & 0.7690 \\
\hline
\end{tabular}
\caption{Vessel segmentation results on the DRIVE dataset}
\label{tab:ds-result}
\end{table}

\begin{table}[t!]
\centering
\begin{tabular}{l|l|l|l|l|l}
\hline
Method & Precision & Recall & F1 Score & Average Max. Dice & Kappa \\
\hline
Deep-FExt & 0.8204 & 0.7954 & 0.8078 & 0.8487 & 0.7920 \\
Deep-FExt (Fusion) & \textbf{0.8299} & \textbf{0.8489} & \textbf{0.8393} & \textbf{0.8699} & \textbf{0.8258} \\
DRIU & 0.8267 & 0.8380 & 0.8323 & 0.8628 & 0.8184 \\
HED & & & 0.805 & & \\
Wavelets & & & 0.774 & & \\
Line Detectors & & & 0.743 & & \\
Second Annotator & 0.6365 & 0.9446 & 0.7605 & 0.7966 & 0.7364 \\
\hline
\end{tabular}
\caption{Vessel segmentation results on the STARE dataset.}
\label{tab:ss-result}
\end{table}

\subsection{Centerline Prediction}
We again test Deep-FExt on DRIVE and STARE datasets for the purpose of centerline prediction. We generated centerline annotations by applying skeletonization to the the various manual annotations and used the same training and testing splits that were used for the vessel segmentation. We evaluated our results based on centerline prediction alone and a combine multi-class prediction of centerline and vessel. We compare these result to the second annotator of these datasets and Deep-FExt outperforms the second annotator (see Tables \ref{tab:dc-result} and \ref{tab:sc-result}).

\begin{table}[t!]
\centering
\begin{tabular}{l|r|r|r|r|r}
\hline
Method & Precision & Recall & F1 Score & Average Max. Dice & Kappa \\
\hline
Deep-FExt (Only Cen)  & 0.5795 & 0.8204 & 0.6792 & 0.7230 & 0.6688 \\
Deep-FExt (Both)  & 0.7138 & 0.7465 & 0.7298 & 0.7720 & 0.7149 \\
Second Annotator (Only Cen) & 0.6038 & 0.4586 & 0.5213 & 0.6395 & 0.4472   \\
Second Annotator (Both) & 0.7045 & 0.6935 & 0.7500 & 0.6989 & 0.6731   \\
\hline
\end{tabular}
\caption{Centerline prediction results on the DRIVE dataset}
\label{tab:dc-result}
\end{table}

\begin{table}[t!]
\centering
\begin{tabular}{l|r|r|r|r|r}
\hline
Method & Precision & Recall & F1 Score & Average Max. Dice & Kappa \\
\hline
Deep-FExt (Only Cen)  & 0.5363 & 0.7427 & 0.6229 & 0.7598 & 0.6145 \\
Deep-FExt (Both)  & 0.7333 & 0.7573 & 0.7451 & 0.7972 & 0.7245 \\
Second Annotator (Only Cen) & 0.5751 & 0.5243 & 0.5485 & 0.6642 & 0.4072  \\
Second Annotator (Both) & 0.6323 & 0.7599 & 0.6902 & 0.7227 & 0.6554   \\
\hline
\end{tabular}
\caption{Centerline prediction results on the STARE dataset}
\label{tab:sc-result}
\end{table}
\section{Conclusion}
Deep-FExt outperforms most of the existing architectures on the DRIVE and STARE datasets. We believe Deep-FExt can be used to extract feature for general medical segmentation tasks. By replacing the 2-D convolutions with 3-D we can also extend Deep-FExt to handle medical volumes. With the idea of mini-networks memory is conserved and speed is also improved. As a further research we recommend experimenting Deep-FExt in an unsupervised manner through auto-encoder fashion. This can help in generating features for clustering in situation where supervised learning is not feasible due to lack of manually annotated dataset. 

\bibliographystyle{splncs03}
\bibliography{references}
\end{document}